%% file: main.tex
\documentclass{article}
\usepackage{iclr2027_conference,times}
\input{math_commands.tex}

\usepackage{amsmath,amssymb,booktabs,float,microtype,url,graphicx,array}
\usepackage[colorlinks=true,allcolors=blue]{hyperref}

\newcommand{\workshoptitle}{Detecting Silent Action-Interface Mismatch\\
with Calibrated Relational World Models}
\title{\workshoptitle}
\input{authors.tex}
\iclrfinalcopy
\hypersetup{pdftitle={Detecting Silent Action-Interface Mismatch with Calibrated Relational World Models},
 pdfauthor={Rui Zhu, Xurui Li, Haixu Tang},pdfsubject={CoRL 2026 UncertainWM Workshop}}

\newcommand{\fail}{\textsc{fail}}
\newcommand{\desc}{\textsc{descriptive}}

\begin{document}
\maketitle
\lhead{CoRL 2026 UncertainWM Workshop}

\begin{abstract}
A robot action can satisfy shape and range constraints yet carry the wrong
semantics: axes may be permuted, signs inverted, rotations rescaled, or gripper
polarity reversed. We study learned state--action--effect relations as offline
contracts for detecting these mismatches and ranking their likely fields. Clean
held-out transitions calibrate alarms; a fixed inverse vocabulary supports
closed-set localization. On ten LIBERO Spatial tasks comprising 432 source
episodes, a capacity-controlled comparison at 2.5\% empirical clean false-positive
rate gives 46.1\% recall for a ridge forward residual, 61.0\% for a four-component
joint Gaussian mixture, and 65.7\% for the exact conditional of that same mixture.
The joint score retains a 14.9 percentage-point advantage over ridge, while
conditioning improves recall by a further 4.7 points. Joint scoring instead has
higher localization: 95.8\% versus 94.6\%. These matched-rate comparisons are
retrospective diagnostics, not deployment calibration guarantees. Earlier
Object/Goal results broaden tasks within the LIBERO convention; DROID, all three
HYDRA tasks, short temporal faults, small offsets, exact repair, and simulated
mitigation fail their respective gates, while xArm contributes one task only.
The evidence supports a bounded offline diagnostic and a detection--localization
tradeoff, rather than joint-density superiority or cross-robot safety.
\end{abstract}

\section{Introduction}

World models usually predict dynamics or support planning. At a controller
boundary, they can also ask whether an action's semantics agree with the observed
transition. A seven-dimensional action vector may mix normalized translation,
axis-angle rotation and gripper commands. Components can agree on shape and bounds
while disagreeing about scale, axes, signs, rotation units or polarity. Such a
command is numerically legal but inconsistent with the recorded state change.

We study whether recorded state--action--next-state tuples can supply an
\emph{executable contract} for this failure class. A detector is trained on clean
transitions, a disjoint clean split fixes its alarm threshold, and in-range semantic
mutations are applied only after that calibration is frozen. After an alarm, the
system evaluates a declared set of inverse transformations and reports the one that
best restores relational consistency. This separates two questions that aggregate
anomaly scores often conflate: whether the transition should be inspected and
which contract field is the most plausible explanation.

The diagnostic addresses interface semantics, not perception, planning, contact or
task progress. Replay supplies the factual next state of the original action, not
the physical outcome of executing a corrupted action. We evaluate offline alarms
and closed-set localization; they do not measure task success or safe correction.

The study separates three sources of apparent advantage: using the observed
transition, representing multimodal clean behavior, and choosing an alarm operating
point. The original joint-mixture comparison outperformed a ridge forward residual,
but those methods had different achieved false-positive rates. A post-decision
capacity-controlled analysis addresses both confounds. At matched empirical clean
FPR, the exact conditional of the fitted mixture has higher detection recall than
the joint score; the joint score retains better closed-set localization. We report
that reversal alongside episode-level uncertainty and the external and closed-loop
failures that bound the method's use.

\section{Relation to prior work}

World models learn compact predictive representations for control
\citep{ha2018worldmodels}; here prediction is used as evidence about an interface,
not as a rollout planner. Learned runtime monitors instead anticipate broad policy
failures. Model-based monitoring uses interactive imitation learning
\citep{liu2023runtime}; Sentinel combines temporal action consistency with task
progress \citep{agia2024sentinel}; and RAPT uses calibrated predictive deviation for
deployment-time detection and diagnosis \citep{munn2026rapt}. These systems address
general failure or distribution shift. Our scope is a declared vocabulary of
controller-semantics transformations and attribution to a contract field.

Model-security work supplies a scope boundary. Backdoor detectability depends on
nominal/hidden-task relation and detector assumptions
\citep{tang2022tasksimilarity,zhu2024gradientshaping}; TrojAI distinguishes
detection, diagnosis and mitigation \citep{reese2026trojai}. Backdoor suppression
\citep{zhu2023selectiveamnesia,jiang2026doubleblind} and model-lifecycle risks
\citep{chen2024janus,wang2025rigging,zhu2026shadowmask} concern different interventions.
They provide no evidence for this fixed-weight interface diagnostic, and inverse
ranking is neither model purification nor adaptive-attack robustness.

Action Mahalanobis distance is a natural cheap out-of-distribution baseline
\citep{lee2018mahalanobis}. It cannot, by construction, use the observed state
change: a transformed action may remain marginally plausible while becoming
inconsistent with the transition. A field-aligned static comparator can couple
translation, rotation, and gripper coordinates to measured displacement, but
public benchmark metadata omit the controller gains and absolute-versus-delta
semantics needed for a faithful executable specification. We include both checks so
that improvement is not measured only against range validation.

Robot-testing work explores a complementary axis. VLATest generates diverse
manipulation scenes \citep{wang2024vlatest}, FATE-VLA searches for failure-prone
scenes \citep{kanwal2026fate}, and metamorphic testing compares related executions
without a task-specific oracle \citep{valle2026metamorphic}. ContractFuzz mutates
action semantics rather than observations or scenes. The motivating deployment
observation is that weights alone do not specify normalization and controller
conventions \citep{tai2026sameweights}; our experiments test when a learned
state--action relation can expose that missing contract.

\section{Executable contract oracle}

\subsection{Threat model and supported vocabulary}

Let $s_t\in\mathbb{R}^{d_s}$, $a_t\in\mathbb{R}^{d_a}$, and $s_{t+1}$ be a clean
recorded transition. A fault $k$ applies a transform
$T_k:\mathbb{R}^{d_a}\rightarrow\mathbb{R}^{d_a}$. We evaluate only changed rows
whose transformed action remains inside clean training bounds,
\begin{equation}
  \mathcal{V}_k=\{t:T_k(a_t)\neq a_t,\;\ell\leq T_k(a_t)\leq u\}.
\end{equation}
The supported set contains halving the six motion coordinates, swapping the X and Y
axes, reversing the X-axis sign, shrinking rotation units, and reversing gripper
polarity. A fixed X-offset is retained as an unsupported stress test. Temporal
delay, half-frequency hold, and chunk reversal are evaluated separately and fail
the main detector gate.

This is a controlled counterfactual on recorded data: the detector receives
$(s_t,T_k(a_t),s_{t+1})$, but $s_{t+1}$ is the recorded outcome of $a_t$. That design
isolates relational inconsistency and supplies a known mutation label. It does not
simulate contact dynamics, policy adaptation, or the task outcome of executing
$T_k(a_t)$.

\subsection{Calibrated relational scores}

From clean training transitions, we fit a ridge forward model
\begin{equation}
f_\theta(s_t,a_t)\approx s_{t+1}-s_t
\end{equation}
and normalize each residual dimension by its training root-mean-square error. The
conditional score is
\begin{equation}
R_f(s_t,a,s_{t+1})=
\sqrt{\operatorname{mean}\left[
\left(\frac{f_\theta(s_t,a)-(s_{t+1}-s_t)}{\sigma_f}\right)^2\right]}.
\end{equation}
We also standardize the joint vector $z=[s_t,a,s_{t+1}-s_t]$. A regularized joint
Gaussian gives
\begin{equation}
R_j(z)=\sqrt{(z-\mu)^\top(\Sigma+10^{-2}I)^\dagger(z-\mu)}.
\end{equation}
The original selected score is the negative log likelihood of a full-covariance
mixture,
\begin{equation}
R_G(z)=-\log\sum_{m=1}^{M}\pi_m
\mathcal{N}(z;\mu_m,\Sigma_m+10^{-3}I),
\end{equation}
with $M=4$ and deterministic split-seed initialization. To control model capacity,
we also evaluate the exact conditional of this fitted mixture. With
$x=[s_t,a]$ and $y=s_{t+1}-s_t$,
\begin{equation}
R_C(y\mid x)=R_G(x,y)+\log p_G(x).
\label{eq:conditional}
\end{equation}
The marginal $p_G(x)$ uses the same weights, component input means and covariance
blocks as $p_G(x,y)$. The conditional mixture therefore has input-dependent
component weights and conditional means, without adding parameters or fitting a
separate model. Its comparison with $R_G$ isolates the contribution of the input
marginal to the score; it does not compare independently optimized model classes.
A disjoint clean calibration
split sets each detector's threshold to an empirical upper quantile targeting 2.5\%
false positives. No mutation outcome enters model selection or threshold fitting.

For attribution, each inverse hypothesis $U_j$, including identity, maps the
observed action back into a candidate convention. The reported candidate is the one
with the lowest relational score on $(s_t,U_j(a),s_{t+1})$. We call Top-1 agreement
with the injected field \emph{localization}; earlier logs use
\texttt{repair\_top1}, but a correct ranking does not demonstrate executable repair.

\section{Frozen evaluation contract}

\paragraph{Data and partitions.}
The primary benchmark is public LIBERO \citep{liu2023libero}, pinned to data revision
\texttt{e5907374380b8f96511957e6ba5582be52a1e179}. Episodes, not transitions, are
partitioned 60/20/20 into training, calibration, and test sets for each deterministic
split seed; at most 2,000 transitions are sampled per seed. The audit contains 1,243
task--episode clusters and 160,414 transitions across 30 retained Spatial, Object,
and Goal tasks. Repeated split seeds reduce algorithmic Monte Carlo variation but do
not create new trajectories.

External-convention screens use low-dimensional data only. DROID-100
\citep{khazatsky2024droid} contributes 100 episodes and 32,112 transitions, but
46/47 task labels are singletons, so its frozen screen is pooled and
episode-disjoint. The FOWM xArm Lift conversion \citep{feng2023finetuning} contains
800 repeated episodes and 19,200 source-semantic XYZ-plus-gripper transitions from
one task. Stanford HYDRA \citep{belkhale2023hydra} contains 570 episodes, three
tasks, and 357,664 valid transitions with source-defined 7D end-effector-delta and
gripper actions. Video and simulator assets are excluded. Conflicting HYDRA mirror
metadata are handled conservatively by retaining the source-specific CC-BY-4.0
notice rather than interpreting license terms.

\paragraph{Metrics, baselines, and gates.}
We report clean false-positive rate (FPR), mutation recall, recall gain over the
named comparator, and closed-set localization Top-1. The cheap baseline is the
better of a full-covariance action Mahalanobis detector and static range legality.
A structured field-aligned displacement model independently fits clean affine maps
from the named action coordinates to next-frame translation, relative axis-angle,
and gripper aperture. The ridge forward score $R_f$ is the original
conditional baseline; the capacity-controlled conditional mixture $R_C$ is added
in the post-decision validation.

\paragraph{Historical gates and uncertainty.}
The original task gate requires FPR at most 0.05, recall gain at least 0.10 over
action/static checks, and localization at least 0.70. Later mixture confirmation
requires a task-median recall gain of 0.10 over ridge, wins on at least eight tasks,
and both methods below 0.05 FPR. These are historical study gates, not matched-FPR
comparisons. The supplement retains the aggregate gates, multiplicity correction,
sensitivity results, and a run whose FPR of 0.0500375 exceeds the limit.
Repeated split seeds are collapsed within source episode before paired task/episode
inference; they do not create new trajectories.

\paragraph{Post-decision validation.}
A fixed protocol evaluates all ten Spatial tasks and five supported families on
100 episode-disjoint partitions per task (seeds 33001--33100), retaining the
original fitting, preprocessing, transition cap and valid-row rule. No new source
episodes are introduced. Each method is first evaluated at clean-calibration
quantiles targeting 1\%, 2.5\%, and 5\% FPR, with achieved test FPR reported
separately. A second, retrospective diagnostic derives thresholds from held-out
clean scores and interpolates adjacent clean-score cutoffs to match empirical FPR
exactly. Mutation labels do not select those cutoffs. This curve compares scores;
it cannot supply a test-free deployment threshold.

The primary endpoint is recall at 2.5\% matched FPR, with two paired comparisons:
joint minus ridge and joint minus conditional GMM. Repeated test appearances are
averaged within episode--fault cells, then faults within episode, episodes within
task, and ten tasks with equal weights. A paired hierarchical bootstrap uses
20,000 task/episode resamples. The reported 95\% intervals are conditional on
fitted scores and clean-reference thresholds; they exclude refitting and
threshold-estimation uncertainty. Bonferroni 97.5\% intervals for the two primary
comparisons are retained in the supplement. Other FPRs and localization are
descriptive endpoints.

\section{LIBERO evidence}

\subsection{Matched FPR and model capacity}

\begin{figure}[t]
\centering
\includegraphics[width=\linewidth]{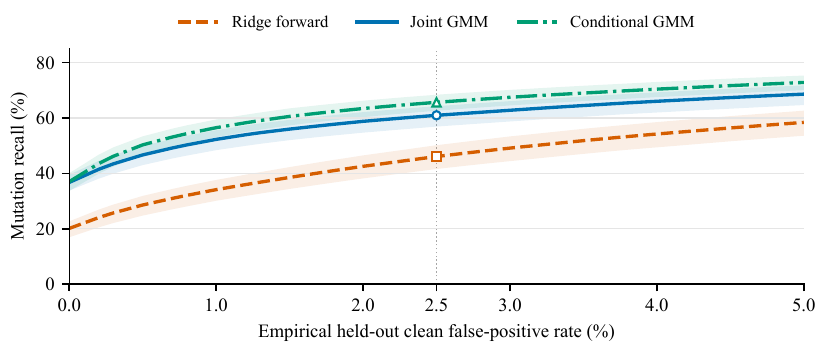}
\caption{Retrospective clean-reference FPR comparison on the same Spatial episodes.
All methods use the same train/calibration/test partitions. Bands are pointwise
95\% task/episode bootstrap intervals conditional on fitted scores and reference
thresholds. They are not simultaneous bands or deployment-FPR guarantees.}
\label{fig:pc-fpr}
\end{figure}

Table~\ref{tab:pc-primary} and Figure~\ref{fig:pc-fpr} compare the three scores
without changing model capacity or empirical clean alarm rate. At 2.5\% FPR, joint
GMM recall is 0.6096 versus 0.4608 for ridge: the paired advantage is 0.1488, with
95\% interval $[0.1207,0.1798]$. The exact conditional GMM instead reaches 0.6566,
exceeding joint recall on all ten tasks. Joint-minus-conditional recall is
$-0.0470$ with interval $[-0.0632,-0.0350]$; its two-comparison 97.5\% interval
$[-0.0658,-0.0339]$ also excludes zero. The conditional advantage is 0.0422 at 1\%
FPR and 0.0422 at 5\%. The result supports multimodal relational modeling over the
ridge residual, but rejects the claim that joint scoring has the best detection
recall among the tested scores.

\begin{table}[t]
\caption{Post-decision Spatial comparison over 432 source episodes. Recall is at
2.5\% matched empirical clean FPR; brackets are conditional paired task/episode
bootstrap 95\% intervals. Achieved FPR uses a separate, calibration-only 2.5\%
nominal threshold and equal-episode weighting. Matched FPR uses clean rows within each split. Localization includes all valid fault rows, not just alarms.}
\label{tab:pc-primary}
\centering\small
\setlength{\tabcolsep}{5pt}
\begin{tabular}{lrrr}
\toprule
Score & Matched recall (\%) & Calibration FPR (\%) & Localization (\%)\\
\midrule
Ridge forward & 46.08 [41.58, 50.17] & 2.63 & 92.70\\
Joint GMM & 60.96 [56.90, 64.28] & 3.22 & 95.81\\
Conditional GMM & 65.66 [62.61, 68.40] & 2.85 & 94.62\\
\bottomrule
\end{tabular}
\end{table}

\subsection{Calibration and localization answer different questions}

At calibration-only thresholds targeting 2.5\% FPR, the achieved equal-episode
clean rates are 0.0263 for ridge, 0.0322 for joint GMM and 0.0285 for conditional
GMM; corresponding recalls are 0.4413, 0.6086 and 0.6504. Sharing a nominal
quantile or satisfying the same 5\% cap therefore does not establish a matched
operating point. The matched analysis removes that confound for this finite
reference set while leaving clean-only deployment calibration as a separate task.

Detection and attribution also rank the models differently. Closed-set localization
is 0.9581 for joint GMM, 0.9462 for conditional GMM and 0.9270 for ridge. These
accuracies use all valid mutated rows under the fixed inverse vocabulary. The
joint score's higher localization does not establish an end-to-end benefit after
alarm selection, and its extra input-marginal term does not improve recall here.

All 1,000 GMM fits converge. Each episode--fault cell has at least 12 held-out
appearances, averaged before inference. The supplement reports independent
conditional-density and exact-FPR interpolation checks.

\subsection{Relation to the retained results}

The original calibration-only comparison reports a task-median joint-minus-ridge
recall difference of 0.15663. A previous 432-episode analysis estimates 0.16753
$[0.13642,0.20297]$. Those values remain in the supplement with their original
operating points; the matched-FPR estimate of 0.1488 above is the appropriate
comparison for the present detection claim. Its different seeds and estimand mean
that subtracting the two estimates does not identify a causal ``FPR contribution.''

Earlier joint-GMM advantages over ridge also recur within LIBERO Object and Goal,
with recall differences 0.21183 and 0.17956 under their original calibration
protocols. They share LIBERO's action convention. The new conditional and
matched-FPR comparisons are restricted to Spatial; they have not been repeated on
those suites. A field-aligned static displacement comparator has median recall
0.32020 and localization 0.84245, below the historical joint-GMM result, but absent
controller gains prevent a faithful manufacturer-specification interpretation.
Complete earlier operating-point, ablation and sensitivity values remain in the
supplement as historical evidence.

\section{Where the contract fails}

\begin{table}[t]
\caption{Material boundaries retained in the main paper. ``Pass'' always refers to
the frozen gate for that row; proxy and simulator rows are not physical trials.}
\label{tab:boundaries}
\centering
\small
\setlength{\tabcolsep}{4.4pt}
\begin{tabular}{p{0.18\linewidth}p{0.14\linewidth}p{0.24\linewidth}p{0.34\linewidth}}
\toprule
Scope & Gate & Result & Supported interpretation\\
\midrule
Small X-offset & 0/10 & GMM recall 0.0340 & Explanation may rank well below alarm threshold\\
Temporal faults & 0/10 & median gain 0.0421 & Short latency/hold/chunk order unsupported\\
DROID-100 & fail & recalls 0.0193--0.0865 & Structural compatibility is not detector transfer\\
xArm Lift & 1 task & GMM recall 0.6016 & One-task offline sanity only\\
HYDRA & 0/3 & recalls 0.1063--0.1213 & Multi-task external convention fails\\
Exact repair & 0/10 & median within-0.01 0.4930 & Localization is not reliable action repair\\
Simulated fallback & fail & ties ridge; GMM FPR misses gate & No ContractFuzz-specific closed-loop gain\\
\bottomrule
\end{tabular}
\end{table}

\subsection{Detection, explanation, and abstention are different decisions}

Fault families vary materially. Supported transformations have median recall gains
0.300--0.452 and localization 0.878--0.966. The fixed X-offset stress test has only
0.0047 gain and 0.555 localization under the original score; the GMM also passes
0/10 offset gates with median recall 0.0340. Yet seven-candidate localization is
0.7340. In a task/episode-clustered mechanism analysis, alarm recall is 0.0328,
localization is 0.7304, and localization among non-alarmed rows is 0.7388; AUROC is
only 0.5475. Relative candidate evidence can therefore be informative even when the
absolute clean-versus-fault distributions overlap too strongly for a high-precision
alarm.

A selection-frozen descriptive reuse of the same 432 episodes places 74.01\% of
offset episodes in an explanation-only false-negative stratum. For supported
families, 75.51\% of episode--fault cells are detected and localized, 1.05\% are
localized without detection, none are joint failures, and 23.43\% lie in a mixed
boundary region. These values explain an existing result; they are not an
independent confirmation.

An omitted-family audit cross-fits confidence thresholds between fixed task halves
using only the other four supported families. Aggregate rejection of a deliberately
hidden family is 0.7411, but omitted X-sign rejection is only 0.3720 and 64.67\% of
those rows are forced into the XY-swap label. This rejects a universal open-set
guarantee. The operational rule is consequently asymmetric: an alarm can request
inspection for a represented family, but no alarm cannot certify a correct
interface, and a high relative score cannot safely remap an unknown fault.

\subsection{External conventions do not validate transfer}

DROID-100 passes low-dimensional schema checks, yet pooled scale, swap, and sign
recalls are only 0.0193, 0.0777, and 0.0865. Its sparse task identity also prevents a
multi-task replication. One xArm Lift task passes a separately frozen screen: GMM
recall is 0.6016 versus 0.5046 for forward, with 0.0245 mean FPR. Because all 800
episodes repeat one task, it supports only a bounded one-task offline sanity check.

HYDRA is the decisive negative external screen. It passes compatibility gates and
provides three repeated tasks with auditable action semantics, but coffee,
dish-rack, and toast GMM recalls are only 0.1063, 0.1097, and 0.1213; zero tasks pass.
Three bounded rescues do not reverse that outcome. The final SAGE-R sequence-aware
redesign also passes 0/3 and underperforms the strongest sequence baseline on every
task. These failures are immutable evidence: the workshop rewrite neither tunes on
the HYDRA test set nor turns the one xArm result into cross-robot validation.

\subsection{Replay evidence does not establish consequence or repair}

Replay uses the factual next state produced by the clean action. A deterministic
clean-trained response proxy lowers median task GMM recall from 0.6379 to 0.2381 on
LIBERO and mean recall from 0.6105 to 0.0972 on xArm, but factual-minus-proxy recall
on HYDRA is $-0.0978/+0.0604/-0.0440$. These learned-response diagnostics are not
environment rollouts or causal effects.

A small controlled simulator connects the detector to reject/hold/fallback under a
fixed fault and delay budget. GMM fallback can recover success relative to no
monitor, but it ties a ridge monitor and misses the GMM clean-FPR gate. It therefore
does not demonstrate ContractFuzz-specific closed-loop success improvement. A
separate continuous-offset repair sweep passes 0/10 gates, and linear, shrunk,
quadratic, random-feature, and disjoint-task selectors all fail. Candidate ranking
is useful diagnostic evidence; it is not a license to modify commands online.

\section{Sensitivity and finite-source limits}

Earlier ablations reject a universal best score. Forward-only and inverse-plus-forward
scores both pass the original action/static gate, but their paired recall
difference includes zero and forward-only localizes better. Four and eight GMM
components pass the historical recall gate while two fail; calibration budgets of
100, 200 and 400 clean transitions pass. Family rankings differ, and small-offset
or temporal severity sweeps cannot rescue the failed short-fault gates. The
supplement retains the full numerical ablations and the distinction between
predeclared comparisons and explanatory reuse.

The new conditional score is an analytic normalization of the same joint mixture.
Its advantage does not establish superiority of every conditional architecture,
and a separately optimized conditional mixture could behave differently. The
post-decision analysis reuses the same 432 episodes, conditions uncertainty on the
fitted scores and reference thresholds, and adds no robot, policy, or controller
convention. Matched empirical FPR describes this finite offline reference set;
production calibration and natural fault prevalence remain outside the evidence.

\section{Implications for uncertain world models}

A multimodal state--action--effect relation can expose declared interface faults
that remain numerically legal. The new comparison changes the interpretation of
that result: a joint-mixture advantage over linear prediction error does not imply
that modeling the joint density is better than conditioning. Here conditioning the
same mixture improves recall, while the joint score ranks inverse candidates more
accurately. Score choice should therefore follow the diagnostic decision and its
alarm budget rather than an aggregate claim of model dominance.

A practical offline audit can pair clean calibration transitions with a declared
inverse vocabulary and request inspection when an alarm occurs. It must report
achieved clean FPR, detection recall and localization separately. Meeting a common
FPR ceiling constrains the operating point but cannot replace a matched-rate
comparison. No alarm cannot certify a correct interface, and candidate ranking
cannot justify automatic command remapping. Independent robot conventions and
actual counterfactual rollouts remain necessary to support broader deployment or
safety claims.

\section{Reproducibility, ethics, and disclosure}

\paragraph{Reproducibility.}
The artifact records pinned data, splits, calibration rules, commands, checksums
and CPU resources. The supplement retains all historical runtime and portability
evidence, including the inconclusive exact-CPython-3.12.4 cross-host replay. The
post-decision analysis adds no source data: 1,000 GMM fits and the conditional/
matched-FPR analysis take 40.40 seconds on one local CPU process with one
BLAS/OpenMP thread (Python 3.11.2, NumPy 1.26.4, SciPy 1.13.1, scikit-learn 1.4.2).
Its protocol, score arrays, episode metrics, hashes and independent aggregation
audit are preserved separately from the original results.

\paragraph{Ethics and scope.}
This diagnostic could be misread as a safety certificate. We publish failures,
restrict the claim to recorded replay, and do not recommend automatic correction.
The evaluation uses public low-dimensional robot demonstrations. Physical
deployment would require independent safety controls and oversight. Dataset license
metadata are provenance facts, not legal conclusions.

\paragraph{AI-use disclosure.}
Generative AI assisted research planning, software development, analysis and
manuscript drafting. This revision used AI assistance for the post-decision code,
mathematical checks, result interpretation, prose and figures. Authors remain
responsible for verifying the evidence, citations and final text.

\section{Conclusion}

On locked LIBERO replay, calibrated relational scores detect and localize five
numerically legal action-interface transformations. At matched empirical clean
FPR, a joint GMM retains its recall advantage over ridge prediction error, but the
exact conditional of that same mixture has still higher recall. Joint scoring
instead has higher closed-set localization. The resulting detection--localization
tradeoff, finite-episode uncertainty and negative external results support an
auditable offline diagnostic for a declared vocabulary. They do not establish
cross-robot generalization, closed-loop success improvement, causal safety or
automatic repair.

\label{workshop-main-end}
\clearpage
\begingroup
\small
\setlength{\bibsep}{0pt plus 0.2ex}
\bibliography{references}
\bibliographystyle{iclr2027_conference}
\endgroup

\clearpage
\begingroup
\setcounter{page}{1}
\setcounter{section}{0}
\setcounter{subsection}{0}
\setcounter{subsubsection}{0}
\setcounter{table}{0}
\setcounter{figure}{0}
\setcounter{equation}{0}
\renewcommand{\theHsection}{supp.\arabic{section}}
\renewcommand{\theHtable}{supp.\arabic{table}}
\renewcommand{\theHfigure}{supp.\arabic{figure}}
\renewcommand{\theHequation}{supp.\arabic{equation}}
\hypersetup{pageanchor=false}
\input{supplement_body.tex}
\endgroup

\end{document}

%% file: math_commands.tex
\usepackage{amsmath,amsfonts,bm}

\def\eqref#1{equation~\ref{#1}}

\def\1{\bm{1}}

\DeclareMathAlphabet{\mathsfit}{\encodingdefault}{\sfdefault}{m}{sl}
\SetMathAlphabet{\mathsfit}{bold}{\encodingdefault}{\sfdefault}{bx}{n}



%% file: authors.tex
\author{Rui Zhu\\
Yale University\\
\texttt{\small rui.zhu.rz399@yale.edu}
\And
Xurui Li\\
Ant Group\\
\texttt{\small xurui.lxr@antgroup.com}
\And
Haixu Tang\\
Indiana University\\
\texttt{\small hatang@iu.edu}}

%% file: supplement_body.tex
\lhead{CoRL 2026 UncertainWM Workshop --- Supplement}
\begin{center}
{\Large\bf Detecting Silent Action-Interface Mismatch with\\
Calibrated Relational World Models}\par
\vspace{0.4em}
{\large Evidence and Reproducibility Supplement}\par
\vspace{0.25em}
{\normalsize Rui Zhu$^1$, Xurui Li$^2$, Haixu Tang$^3$}\par
{\small $^1$Yale University \quad $^2$Ant Group \quad $^3$Indiana University}\par
{\footnotesize\texttt{rui.zhu.rz399@yale.edu}, \texttt{xurui.lxr@antgroup.com}, \texttt{hatang@iu.edu}}
\end{center}
\vspace{0.5em}

\section{Scope and evidence hierarchy}

This supplement separates the retained original study from the post-decision
matched-FPR and conditional-score validation in Section~\ref{sec:pc-validation}.
The latter reuses existing source episodes and introduces new scoring/operating-point
results; all other reported values retain their original evaluation contracts and
are linked to project-local commands and machine-readable results in
Section~\ref{sec:repro}. The positive claim remains restricted to \emph{offline
replay of five declared, numerically legal action-interface transformations within
the LIBERO convention}. DROID and HYDRA remain negative; xArm remains a one-task
sanity check; simulator mitigation, exact repair, open-set rejection, small offsets,
and short temporal faults remain unsupported.

Table~\ref{tab:evidence-levels} defines how each evidence level may affect the
paper. Repeated split seeds characterize an algorithm under different
episode-disjoint partitions; they are not independent robot executions. The
episode-clustered analysis is the inferential anchor. Analyses that reuse those
episodes after inspecting earlier outcomes are explicitly descriptive. External
screens probe convention transfer but do not become positive replications unless
their predeclared gate passes.

\begin{table}[H]
\caption{Evidence hierarchy. ``May support'' is deliberately narrower than what
the underlying metric could be used to suggest.}
\label{tab:evidence-levels}
\centering
\small
\setlength{\tabcolsep}{4.5pt}
\begin{tabular}{p{0.19\linewidth}p{0.31\linewidth}p{0.40\linewidth}}
\toprule
Level & Statistical unit & May support \\
\midrule
Fresh locked replay & task summaries over unseen split seeds & Detector ordering and clean-FPR compliance on recorded tasks \\
Clustered confirmation & task, then immutable source episode & Finite-source Spatial effect and paired uncertainty \\
Selection-frozen reuse & previously used task--episodes & Mechanism, heterogeneity, and failure description only \\
Post-hoc severity & task summaries at fixed severities & Explanatory sensitivity, not rescue of a failed family \\
External convention & source-defined episodes and tasks & Transfer only when the independently frozen gate passes \\
Simulator proxy & simulator episodes under matched budgets & Only the specified mitigation comparison, never physical safety \\
\bottomrule
\end{tabular}
\end{table}

\paragraph{Decision vocabulary.}
An alarm is a thresholded clean-versus-mutated decision. Localization is Top-1
ranking among a fixed inverse-candidate set after an alarm score has been computed.
Open-set rejection asks whether a held-out family should be rejected rather than
forced into a known label. Repair asks whether an inferred continuous magnitude is
accurate enough to modify an action. These are four different estimands. The paper
supports the first two for represented LIBERO families and does not collapse them
into a claim about open-set coverage, action correction, or task success.

\section{Frozen data and transformation contract}
\label{sec:contract}

\subsection{Source units and episode firewall}

LIBERO is pinned to revision
\texttt{e5907374380b8f96511957e6ba5582be52a1e179}. Table~\ref{tab:data} reports
the source units before any repeated split generation. Each deterministic split
assigns whole episodes 60/20/20 to train, clean calibration, and test. At most 2,000
transitions are sampled per seed. Calibration uses an empirical upper quantile with
\texttt{method=higher}; mutation labels and outcomes are unavailable to fitting,
model selection, and threshold choice.

\begin{table}[H]
\caption{Immutable source units. A task--episode is the inferential cluster; a
transition is not treated as an independent trial.}
\label{tab:data}
\centering
\small
\begin{tabular}{lrrrr}
\toprule
Suite & Tasks & Transitions & Task--episodes & Episodes/task \\
\midrule
LIBERO Spatial & 10 & 52,538 & 432 & 35--47 \\
LIBERO Object & 10 & 56,262 & 383 & 31--41 \\
LIBERO Goal & 10 & 51,614 & 428 & 33--50 \\
\midrule
Total & 30 & 160,414 & 1,243 & 31--50 \\
\bottomrule
\end{tabular}
\end{table}

An independent coverage audit generates 3,000 episode-disjoint partitions with
1,200/400/400 sampled train/calibration/test transitions. Every source episode
appears in every role across the audit, with minimum assignment counts 40, 9, and
10; each task has 100 unique role assignments. These checks establish split
coverage, not additional data breadth.

\subsection{Supported vocabulary and valid-row rule}

The seven-dimensional action contains six motion coordinates and one gripper
coordinate. Table~\ref{tab:families} fixes the five supported hypotheses and the
unsupported offset control. A row is valid only if the transformation changes its
action and every transformed coordinate stays inside the clean-training bounds.
Scoring uses every valid corrupted test row; it does not retain only examples that
are easy to classify. The inverse-candidate set contains identity and one inverse
per declared family.

\begin{table}[H]
\caption{Frozen mutation vocabulary. Motion transformations act on the first six
coordinates unless a smaller coordinate set is named.}
\label{tab:families}
\centering
\small
\begin{tabular}{lll}
\toprule
Status & Family & Exact transformation \\
\midrule
Supported & Motion scale & $a_{1:6}\leftarrow 0.5a_{1:6}$ \\
Supported & XY permutation & $(a_x,a_y)\leftarrow(a_y,a_x)$ \\
Supported & X sign & $a_x\leftarrow-a_x$ \\
Supported & Rotation units & $a_{4:6}\leftarrow a_{4:6}/57.2957795$ \\
Supported & Gripper polarity & $a_7\leftarrow 1-a_7$ \\
Stress test & Fixed X offset & $a_x\leftarrow a_x+0.05$ \\
\bottomrule
\end{tabular}
\end{table}

The locked per-task gate requires clean FPR $\leq0.05$, recall gain over the
action/static comparator $\geq0.10$, and localization $\geq0.70$. The Spatial
aggregate requires at least 8/10 task passes, median gain $\geq0.15$, median
localization $\geq0.75$, and maximum task FPR $\leq0.05$. The later GMM comparison
requires GMM-minus-forward median recall $\geq0.10$, wins on at least 8/10 tasks,
and both maximum task FPRs $\leq0.05$. No rounded value is used for a gate.

\section{Task-level mixture confirmation}
\label{sec:task-results}

Table~\ref{tab:gmm-task} exposes all ten primary Spatial task summaries rather than
only their median. The recall comparison uses 1,000 fresh episode-split seeds
(21971--22970). The localization columns use a separate 500-seed confirmation. GMM
recall exceeds forward on 10/10 tasks; the median gain is 0.15663 and maximum GMM
clean FPR is 0.03994. Median GMM localization is 0.95926, with median advantages
0.02932 over forward and 0.01728 over the joint Gaussian. One convergence warning
occurs among 5,000 fits and does not alter the frozen result.

\begin{table}[H]
\caption{Fresh Spatial task results for the four-component GMM. $\Delta R$ is
GMM-minus-forward recall; $\Delta L_f$ and $\Delta L_j$ are localization gains over
forward and joint Gaussian.}
\label{tab:gmm-task}
\centering
\small
\setlength{\tabcolsep}{4.8pt}
\begin{tabular}{rrrrr}
\toprule
Task & $\Delta R$ & GMM loc. & $\Delta L_f$ & $\Delta L_j$ \\
\midrule
0 & 0.1605 & 0.9632 & 0.0281 & 0.0151 \\
1 & 0.2431 & 0.9656 & 0.0232 & 0.0172 \\
2 & 0.1555 & 0.9577 & 0.0310 & 0.0181 \\
3 & 0.2653 & 0.9466 & 0.0414 & 0.0287 \\
4 & 0.0826 & 0.9417 & 0.0305 & 0.0197 \\
5 & 0.1035 & 0.9650 & 0.0315 & 0.0129 \\
6 & 0.1527 & 0.9530 & 0.0247 & 0.0148 \\
7 & 0.1772 & 0.9561 & 0.0275 & 0.0173 \\
8 & 0.1421 & 0.9642 & 0.0401 & 0.0197 \\
9 & 0.1578 & 0.9608 & 0.0243 & 0.0127 \\
\midrule
Median & 0.1566 & 0.9593 & 0.0293 & 0.0173 \\
\bottomrule
\end{tabular}
\end{table}

The first 200-seed GMM screen is retained separately: recall wins on 10/10 tasks,
but task 4 has clean FPR 0.0500375, exceeding 0.05 by 0.0000375. Its classification
is \emph{inconclusive/near-boundary}, not pass. The fresh 1,000-seed run was
predeclared before its outcomes were inspected and passes without relying on
rounding.

The frozen extension within LIBERO changes task content but not action convention.
On Object, GMM-minus-forward recall is 0.21183 with task-bootstrap interval
$[0.2016,0.2218]$; GMM localization is 0.9521, advantage 0.02888, and maximum FPR
0.0346. On Goal, the corresponding values are 0.17956 $[0.1743,0.1849]$, 0.9525,
0.02758, and 0.0288. These are within-convention suite confirmations, not
independent robot evidence.

\section{Episode-clustered inference and multiplicity}
\label{sec:clustered}

The predeclared estimand first averages repeated test appearances for each
episode--fault cell, then averages faults within episode and episodes within task,
and finally assigns equal weight to each of ten Spatial tasks. The paired
hierarchical bootstrap resamples tasks and then source episodes within task for
100,000 draws. Split-randomization replicates are collapsed and are never counted
as independent clusters.

The family contains exactly three directional comparisons: GMM-minus-forward
recall, GMM-minus-forward localization, and GMM-minus-joint localization.
Table~\ref{tab:clustered} reports ordinary paired 95\% intervals, one-sided
family-wise lower bounds using $\alpha/3$, and exact task-sign-flip tests with Holm
correction. All three lower bounds are positive. Each immutable episode appears in
at least 11 test splits; all 1,000 task-level GMM fits converge; maximum task-mean
clean FPR is 0.0444.

\begin{table}[H]
\caption{Fresh 432-episode Spatial effects. The sign-flip family has ten paired
task summaries, so its smallest attainable unadjusted one-sided value is 1/1024.}
\label{tab:clustered}
\centering
\small
\setlength{\tabcolsep}{3.8pt}
\begin{tabular}{lrrrc}
\toprule
Effect & Estimate & Paired 95\% CI & FWER lower & Holm $p$ \\
\midrule
GMM $-$ forward recall & 0.16753 & [0.13642, 0.20297] & 0.13401 & 0.00293 \\
GMM $-$ forward loc. & 0.03089 & [0.02679, 0.03541] & 0.02645 & 0.00293 \\
GMM $-$ joint loc. & 0.01837 & [0.01522, 0.02198] & 0.01497 & 0.00293 \\
\bottomrule
\end{tabular}
\end{table}

An unchanged attempt to extend this clustered protocol to all suites aborts before
inference because Object task 5, split seed 25221, contains zero valid
gripper-polarity test rows under the locked valid-row rule. The evaluator is not
altered after that observation. Consequently, no episode-clustered Object/Goal
claim is made; only the earlier locked task-bootstrap suite confirmation is
reported.

\section{Comparators, families, and sensitivity}
\label{sec:sensitivity}

\subsection{Why the mixture was selected in the original study}

In a fresh Gaussian comparison, the conditional forward score has higher recall
than a joint Gaussian on every Spatial task (median advantage 0.1033), while its
localization is lower on every task (median forward-minus-joint $-0.0118$). The
predeclared dominance gate passes only 6/10 tasks and therefore fails. This is a
real falsification of the earlier forward-dominance hypothesis, not an ablation
removed after the mixture became favorable.

The tradeoff is family-specific (Table~\ref{tab:gaussian-family}). Forward wins
both metrics for scale, joint wins both for X-sign, and only rotation satisfies the
material Pareto gate. Aggregate rankings must therefore be read as decisions for a
fixed family mixture rather than universal detector dominance.

\begin{table}[H]
\caption{Forward-minus-joint Gaussian effects by transformation family. Positive
localization favors forward; negative localization favors joint.}
\label{tab:gaussian-family}
\centering
\small
\begin{tabular}{lrrl}
\toprule
Family & Recall effect & Loc. effect & Interpretation \\
\midrule
Scale & 0.2620 & 0.0098 & Forward wins both \\
XY permutation & 0.0353 & -0.0248 & Recall--location tradeoff \\
X sign & -0.0242 & -0.0277 & Joint wins both \\
Rotation units & 0.0990 & -0.0168 & Only family passing Pareto gate \\
Gripper polarity & 0.1054 & -0.0036 & Recall--location tradeoff \\
\bottomrule
\end{tabular}
\end{table}

The historical static comparator fits clean affine maps from named action fields to
next-frame translation, relative axis-angle motion, and gripper aperture. Its
task-median recall is 0.32020, localization 0.84245, and maximum FPR 0.029825, but
it passes 0/10 noninferiority gates against GMM. Metadata-minus-GMM recall is
$-0.31196$ with paired interval $[-0.33328,-0.24729]$; localization is $-0.10914$
$[-0.12655,-0.09866]$. The source metadata do not expose controller gains and
absolute-versus-delta semantics, so this is a faithful field-aligned empirical
comparator, not a claimed implementation of a complete manufacturer contract.

\subsection{Mixture, calibration, and operating-point sensitivity}

Table~\ref{tab:gmm-sensitivity} reports the two frozen one-factor sensitivity
families. Four and eight components pass; two components fail the recall-advantage
gate. Nested clean calibration caps all pass. Neither result justifies arbitrary
mixture-count invariance or eliminating the held-out calibration split.

\begin{table}[H]
\caption{GMM sensitivity. Intervals apply to GMM-minus-forward recall; FPR is the
largest task value.}
\label{tab:gmm-sensitivity}
\centering
\small
\setlength{\tabcolsep}{4.3pt}
\begin{tabular}{llrrrl}
\toprule
Factor & Setting & Recall gain & 95\% interval & Loc. & Max FPR \\
\midrule
Components & 2 & 0.0725 & [0.0407,0.1149] & 0.9561 & 0.0443 (fail) \\
 & 4 & 0.1609 & [0.1264,0.2049] & 0.9585 & 0.0458 \\
 & 8 & 0.1634 & [0.1274,0.2118] & 0.9566 & 0.0424 \\
\midrule
Calibration cap & 100 & 0.1453 & [0.1194,0.1892] & 0.9607 & 0.0412 \\
 & 200 & 0.1540 & [0.1269,0.1983] & 0.9607 & 0.0353 \\
 & 400 & 0.1482 & [0.1271,0.2024] & 0.9607 & 0.0362 \\
\bottomrule
\end{tabular}
\end{table}

The earlier forward OFAT audit shows the same need to bind claims to an operating
point. The locked baseline has gain 0.3763, localization 0.9297, and FPR 0.0314.
Budgets of 500 and 1,000 preserve the ordering (gains 0.3639 and 0.3678). A 1\%
target lowers gain to 0.2858 and FPR to 0.0162. A nominal 5\% target raises gain to
0.4389 but produces FPR 0.0601 and fails. Ridge values $10^{-4}$ and $10^{-2}$ are
identical at reported precision. Mild versus severe transformations yield gains
0.2790 and 0.4362. Threshold and severity are therefore experimental factors, not
post-hoc knobs.

\section{Failure analyses that bound the claim}
\label{sec:failures}

\subsection{Unsupported offset, temporal faults, and open set}

The fixed $+0.05$ X offset is an intentionally unsupported family. In the fresh
episode-clustered audit, alarm recall is 0.03282, localization 0.73042,
localization among non-alarmed rows is 0.73881, and alarm AUROC is 0.54746. Thus a
relative candidate can rank correctly even when the absolute score cannot separate
clean from corrupted rows. A selection-frozen reuse places 74.01\% of episodes in
an explanation-only false-negative stratum. This mechanism result is
\desc{}, not independent confirmation.

For supported families, a descriptive reuse reports 75.51\% detected-and-localized
episode--fault cells, 1.05\% localized without detection, 0\% joint failures, and
23.43\% mixed boundary cells. The true candidate is Top-1 on 95.83\% of supported
cells; identity confusion is 2.685\%. These proportions explain where the closed-set
score succeeds and do not add source episodes.

In a cross-fitted leave-one-family-out audit, known-family acceptance is 0.90007 and
unknown-family rejection is 0.74115 (Holm $p=0.001953$). The aggregate hides a
material failure: omitted X-sign rejection is only 0.37200, and 64.67\% of its rows
are forced into the XY label. The system therefore has no universal open-set
guarantee.

Short temporal faults also fail: forward-minus-action gains are 0.0055 for
one-frame delay, 0.0364 for two-frame delay, 0.0061 for half-frequency hold, and
0.1277 for chunk reversal; the all-family median is 0.0421 and 0/10 tasks pass. A
clean-trained temporal feature model reaches recalls 0.0253/0.0247/0.0000 for the
first three families and is worse than the strongest static comparator by
0.0113/0.0391/0.0373. A post-hoc delay severity curve rises sharply at four, eight,
and sixteen frames (gains 0.2926/0.7179/0.9209), but it explains rather than rescues
the frozen short-delay failure.

\subsection{External conventions and bounded redesigns}

Table~\ref{tab:external} contains every independent-convention outcome. DROID's
structural gate passes, but sparse task identity (46/47 labels are singletons)
limits the frozen analysis to pooled, episode-disjoint screens. Only one xArm Lift
task is available, so its positive result cannot support multi-task or cross-robot
generalization. HYDRA supplies three auditable repeated tasks and is the decisive
negative screen.

\begin{table}[H]
\caption{External-convention outcomes. All use low-dimensional source-semantic
actions; video and simulator assets are excluded.}
\label{tab:external}
\centering
\small
\setlength{\tabcolsep}{3.5pt}
\begin{tabular}{p{0.14\linewidth}p{0.18\linewidth}p{0.35\linewidth}p{0.22\linewidth}}
\toprule
Source & Units & Detector result & Classification \\
\midrule
DROID-100 & 100 ep.; pooled & scale/swap/sign recall 0.019/0.078/0.087 & \fail{} transfer; sparse identity \\
xArm Lift & 800 ep.; 1 task & GMM/forward recall 0.602/0.505; loc. 0.929; FPR 0.024 & One-task offline sanity \\
HYDRA & 570 ep.; 3 tasks & GMM recall 0.106/0.110/0.121; loc. 0.728--0.792 & 0/3, \fail{} transfer \\
\bottomrule
\end{tabular}
\end{table}

Three bounded, predeclared rescue candidates fail to turn any HYDRA task positive.
A subsequent nontrivial SAGE-R sequence-aware redesign also passes 0/3 and
underperforms the strongest sequence baseline on all tasks: its attribution values
are 0.3000/0.2576/0.1111 and its gains over that baseline are
$-0.2933/-0.0909/-0.3175$. These failures terminated the ICLR expansion. The
workshop retarget preserves them as historical evidence and does not reinterpret
the rewrite as another rescue candidate.

\subsection{Factual replay is not counterfactual consequence}

In replay, the next state is the factual outcome of the original clean action, not
the outcome that the transformed command would have produced. A deterministic
clean-trained response proxy reduces median GMM recall from 0.6379 to 0.2381 on
LIBERO and mean recall from 0.6105 to 0.0972 on xArm, but HYDRA factual-minus-proxy
effects are $-0.0978/+0.0604/-0.0440$. This inconsistency prevents a causal reading.

The smallest controlled simulator compares no monitor, ridge, GMM, and oracle under
the same injected faults and response-delay budget. Ridge and GMM both improve
success from 0.333 to 1.000, so the ContractFuzz-specific difference is exactly
zero; GMM clean FPR is 0.08 and misses the 0.05 gate. Continuous-offset repair also
passes 0/10 gates (median within-0.01 accuracy 0.4930). These results support
neither a detector-specific closed-loop benefit nor automatic action correction.

\section{Compute, artifacts, and exact reproduction}
\label{sec:repro}

\subsection{Main-evidence resource accounting}

All retained scientific evidence is CPU-only. Table~\ref{tab:compute} reports
full-workload wall time and exact maximum resident set size (RSS) for the ten
headline paths after schema-aware replay. Historical runtimes sum to 959.860
seconds. RSS is process maximum, not machine reservation; these runs were executed
one at a time.

\begin{table}[H]
\caption{Exact CPU resource evidence for the ten retained main-evidence paths.}
\label{tab:compute}
\centering
\small
\setlength{\tabcolsep}{4.1pt}
\begin{tabular}{llrr}
\toprule
ID & Role & Wall (s) & Max RSS (bytes) \\
\midrule
042b & fresh GMM recall & 465.445 & 220,004,352 \\
043 & fresh localization & 225.377 & 250,036,224 \\
044 & Object/Goal GMM & 43.810 & 960,430,080 \\
051 & static metadata baseline & 51.542 & 241,270,784 \\
053a & split coverage & 0.899 & 42,762,240 \\
054a & clustered effects & 46.422 & 588,038,144 \\
055 & offset failure & 29.123 & 752,975,872 \\
056 & false-negative strata & 1.765 & 554,205,184 \\
057 & supported heterogeneity & 32.804 & 760,741,888 \\
058 & open-set audit & 62.673 & 482,066,432 \\
\bottomrule
\end{tabular}
\end{table}

A relocation audit copied an existing environment prefix and reproduced the
artifact in 150.34 seconds with 593,002,496 bytes maximum RSS; the relocated prefix
occupies 4,345,249,565 bytes. It establishes path relocation of the captured
environment, not an independent dependency installation. Six core tests,
synthetic smoke, archive unpack/checksum, warning checks, and final consistency pass
under that interpretation. A separate CPython 3.12.3 host passes the tests but does
not satisfy an exact-3.12.4 contract. The exact CPython 3.12.4 standalone tail is
inconclusive: its first harness had a parser defect, and the authorized one-shot
continuation ended at DNS transport before SSH. We therefore make no exact-patch
cross-host portability claim.

The original main-text summary additionally records the 4,000-seed Spatial and
Object/Goal stages at 499.45 and 96.24 seconds, respectively, and a full replay of
the original clustered analysis at 48.83 seconds with 588,038,144 bytes maximum
RSS. These historical measurements are distinct from the new 40.40-second local
validation and are not added as independent scientific evidence.

\subsection{Command-to-result locator}

Table~\ref{tab:locator} is a compact human index. The project-local file
\path{artifact/table-command-map.md} contains all 31 table rows with exact commands;
\path{artifact/evidence-ledger-052.tsv} is the append-only disposition ledger.
Result JSONs store the complete numeric payload, contract, seed range, and gate
classification. Running a command is not permission to replace a frozen result:
reproduction writes a new output path and a schema-aware verifier compares the
scientific payload.

\begin{table}[H]
\caption{Key evidence locators. Paths are relative to the private project root.}
\label{tab:locator}
\centering
\scriptsize
\setlength{\tabcolsep}{3pt}
\begin{tabular}{lll}
\toprule
Claim & Command/script locator & Result locator \\
\midrule
GMM recall & \texttt{run-gmm-confirmation-042b.sh} & \texttt{gmm-joint-confirmation-042b.json} \\
GMM localization & \texttt{run-gmm-localization-043.sh} & \texttt{gmm-localization-confirmation-043.json} \\
Suite repeat & \texttt{run-cross-suite-gmm-044.sh} & \texttt{gmm-cross-suite-confirmation-044.json} \\
Static comparator & \texttt{run-metadata-displacement-baseline-051.sh} & \texttt{metadata-displacement-baseline-051.json} \\
Split coverage & \texttt{run-statistical-unit-audit-053a.sh} & \texttt{statistical-unit-audit-053a.json} \\
Clustered effects & \texttt{run-clustered-effect-intervals-054a.sh} & \texttt{clustered-effect-intervals-054a.json} \\
Offset failure & \texttt{run-offset-episode-mechanism-055.sh} & \texttt{offset-episode-mechanism-055.json} \\
Failure strata & \texttt{run-offset-episode-trajectories-056.sh} & \texttt{offset-episode-trajectories-056.json} \\
Heterogeneity & \texttt{run-supported-family-episode-traces-057.sh} & \texttt{supported-family-episode-traces-057.json} \\
Open-set audit & \texttt{run-open-set-candidate-rejection-058.sh} & \texttt{open-set-candidate-rejection-058.json} \\
DROID screen & \texttt{run-droid100-sanity-069b.sh} & \texttt{droid100-cross-convention-sanity-069b.json} \\
xArm screen & \texttt{run-xarm-lift-sanity-077b.sh} & \texttt{xarm-lift-cross-convention-sanity-077b.json} \\
HYDRA screen & \texttt{run-stanford-hydra-sanity-080b.sh} & \texttt{stanford-hydra-cross-task-sanity-080b.json} \\
\bottomrule
\end{tabular}
\end{table}

The short names above are reading aids, not substitutes for the exact map; the audit
verifies each named result against the actual registered path. The original review
artifact excludes raw demonstrations, videos, simulator assets, caches, model
weights, credentials, hostnames, home-directory paths, and author metadata. It
includes evaluation code, tests, small aggregate results, manifests, licenses and
notices, command maps, and the frozen split/evaluator contract.

\subsection{Minimal reproduction order}

The fail-closed order is: (1) verify archive SHA-256 and unpack into a fresh local
directory; (2) verify the dependency manifest and data revision/checksum manifest;
(3) run the bundled core test suite; (4) run the data-free synthetic smoke; (5)
execute a chosen evidence command with a new output path and fixed seed contract;
(6) compare the resulting scientific payload with its registered JSON using the
schema-aware verifier; and (7) record wall time, RSS, warnings, and the final
consistency result. Missing data or dependency evidence is a reproduction failure,
not permission to fetch a different revision or relax the evaluator.

\section{Disclosure and reviewer checklist}

\paragraph{Authorship and disclosure.}
This camera-ready uses the author-confirmed names, order, affiliations and emails.
Generative AI assisted planning, software development, analysis and manuscript
drafting, including the post-decision validation and revision. Authors remain
responsible for the evidence, citations and final text. License entries report
provenance and are not legal conclusions.

\paragraph{Reviewer-verifiable boundaries.}
A reviewer should be able to answer yes to each of the following directly from the
paper and supplement:
\begin{enumerate}\small
  \setlength{\itemsep}{0pt}\setlength{\parsep}{0pt}\setlength{\topsep}{2pt}
  \item Episode-disjoint fitting; clean calibration separated from retrospective reference thresholds?
  \emph{Yes.}
  \item Positive unit is LIBERO tasks/episodes, not split seeds? \emph{Yes.}
  \item Matched-FPR results and the conditional-GMM reversal visible for all ten tasks? \emph{Yes.}
  \item Multiplicity and failed all-suite cluster extension disclosed? \emph{Yes.}
  \item Near-boundary and detector/family/repair/mitigation failures retained?
  \emph{Yes.}
  \item DROID/HYDRA negative; xArm one-task-only? \emph{Yes.}
  \item Runtime, RSS, locators, and portability limitation available? \emph{Yes.}
  \item No cross-robot, physical closed-loop, causal-safety, or repair claim?
  \emph{Yes.}
\end{enumerate}

\paragraph{What would change the scientific conclusion.}
Widening requires new evidence: multiple independent tasks under a second auditable
action convention, predeclared advantage over the strongest sequence baseline, and
a real counterfactual environment where mitigation improves task success under the
same fault and delay budget. These nontrivial extensions are outside this rewrite;
more split seeds, post-failure threshold changes, or descriptive trajectory
selection cannot substitute for them.

\clearpage
\input{pc_validation_supplement.tex}

\input{retained_numerical_details.tex}

\label{workshop-supplement-end}

%% file: pc_validation_supplement.tex
\section{Post-decision matched-FPR and capacity-controlled validation}
\label{sec:pc-validation}

\subsection{Protocol, data reuse, and exact conditional score}

This analysis was specified after the PC decision and before observing its new
outputs. It reuses all 432 original Spatial source episodes and is not an independent
replication. Ten tasks, five supported fault families, the original valid-row rule,
training standardization, ridge regularization and four-component full-covariance
GMM are unchanged. Each task uses 100 episode-disjoint 60/20/20 partitions with
seeds 33001--33100 and the original 2,000-transition cap. All 1,000 GMM fits converge;
each episode--fault cell has at least 12 held-out appearances. No mutation outcome
selects a seed, component count, threshold or reported family.

For standardized $x=(s,a)$ and $y=s'-s$, let $p_G(x,y)$ be the already fitted joint
GMM. Its input marginal uses the same weights and covariance blocks:
\[
 p_G(x)=\sum_m\pi_m\mathcal N(x;\mu_{x,m},\Sigma_{xx,m}),\qquad
 R_C=-\log p_G(x,y)+\log p_G(x).
\]
Equivalently, the conditional mixture has weights proportional to
$\pi_m\mathcal N(x;\mu_{x,m},\Sigma_{xx,m})$, conditional component means
$\mu_{y,m}+\Sigma_{yx,m}\Sigma_{xx,m}^{-1}(x-\mu_{x,m})$, and covariance
$\Sigma_{yy,m}-\Sigma_{yx,m}\Sigma_{xx,m}^{-1}\Sigma_{xy,m}$.
Thus joint and conditional scores have identical fitted capacity and parameters.
This is a scoring ablation; it does not optimize a separate conditional likelihood.
An independent block-conditional calculation agrees to $2.3\times10^{-15}$.

\subsection{Clean-reference FPR and deployment calibration}

Calibration-only thresholds use the original upper empirical clean-calibration
quantile with \texttt{method=higher}. Achieved test FPR is reported, not assumed
from the nominal target. Separately, each fitted score is evaluated against its
held-out clean reference: adjacent clean-score cutoffs are interpolated to give
exact empirical FPR at the specified targets. The interpolation coefficients use
clean scores only; no mutant score chooses a cutoff. The maximum numerical FPR
error is $1.2\times10^{-16}$, including tied-score self-checks.

Matched FPR is transition-weighted within each held-out clean split. Mutation
recall averages within episode--fault cells across repeated held-out appearances,
then equally over faults, episodes and tasks. Calibration-only test FPR in
Table~\ref{tab:pc-calibration} uses that equal-episode aggregation. These weighting
choices are explicit: a 2.5\% matched clean-reference rate is not a claim that each
episode individually has a 2.5\% alarm rate. Reference thresholds are used only for
retrospective score comparison, not proposed as deployable thresholds.

\begin{table}[H]
\caption{Matched empirical clean FPR: mutation recall in percent. All rows use the
same finite source episodes, so different FPRs are correlated descriptive views.}
\label{tab:pc-all-fpr}
\centering\small
\begin{tabular}{rrrr}
\toprule
FPR (\%) & Ridge forward & Joint GMM & Conditional GMM\\
\midrule
1.00 & 34.13 & 52.27 & 56.49\\
2.50 & 46.08 & 60.96 & 65.66\\
5.00 & 58.41 & 68.62 & 72.83\\
\bottomrule
\end{tabular}
\end{table}

\begin{table}[H]
\caption{Calibration-only operating points in percent. Each cell gives achieved
clean FPR / mutation recall; calibration targets are nominal, not achieved rates.}
\label{tab:pc-calibration}
\centering\small
\begin{tabular}{lrrr}
\toprule
Score & Nominal 1\% & Nominal 2.5\% & Nominal 5\%\\
\midrule
Ridge forward & 1.16 / 31.34 & 2.63 / 44.13 & 5.05 / 57.26\\
Joint GMM & 1.71 / 51.09 & 3.22 / 60.86 & 5.62 / 68.50\\
Conditional GMM & 1.34 / 54.76 & 2.85 / 65.04 & 5.32 / 72.64\\
\bottomrule
\end{tabular}
\end{table}

\subsection{Paired uncertainty and endpoints}

For the primary 2.5\%-FPR endpoint, the two declared comparisons are joint minus
ridge and joint minus conditional. Repeated split appearances are collapsed before
20,000 paired hierarchical task/episode bootstrap draws (seed 20261017). Ordinary
95\% intervals and two-sided 97.5\% intervals for the two-comparison Bonferroni
family are shown below. These intervals are conditional on fitted models and clean
reference thresholds. They do not propagate training refits or reference-threshold
estimation, and cannot establish population-level deployment-FPR control.

\begin{table}[H]
\caption{Primary matched-FPR paired effects, in percentage points. The wider
intervals control the two declared comparisons by Bonferroni under the stated
conditional bootstrap interpretation.}
\label{tab:pc-effects}
\centering\small
\begin{tabular}{lrrr}
\toprule
Comparison & Effect & 95\% interval & 97.5\% interval\\
\midrule
Joint $-$ ridge & 14.88 & [12.07, 17.98] & [11.72, 18.43]\\
Joint $-$ conditional & -4.70 & [-6.32, -3.50] & [-6.58, -3.39]\\
\bottomrule
\end{tabular}
\end{table}

\begin{table}[H]
\caption{Task-level recall at matched 2.5\% empirical clean FPR, in percent.
Task means equal-weight their source episodes and five represented families.}
\label{tab:pc-tasks}
\centering\small
\begin{tabular}{rrrrr}
\toprule
Spatial task & Episodes & Ridge & Joint & Conditional\\
\midrule
0 & 46 & 47.98 & 62.57 & 67.76\\
1 & 42 & 45.12 & 66.84 & 69.45\\
2 & 39 & 45.59 & 61.01 & 65.07\\
3 & 35 & 30.80 & 54.84 & 57.54\\
4 & 45 & 39.22 & 46.67 & 57.66\\
5 & 43 & 58.18 & 67.80 & 72.93\\
6 & 47 & 46.26 & 60.49 & 65.23\\
7 & 45 & 47.87 & 63.86 & 66.87\\
8 & 46 & 50.85 & 64.07 & 68.52\\
9 & 44 & 48.97 & 61.50 & 65.60\\
\bottomrule
\end{tabular}
\end{table}

The conditional score has higher detection recall than joint on all ten tasks.
Joint nevertheless has higher overall closed-set localization: 95.81\%
$[95.23,96.31]$, versus 94.62\% $[94.07,95.12]$ for conditional and 92.70\%
$[91.93,93.40]$ for ridge (descriptive 95\% intervals). Localization is calculated
over all valid fault rows. It is neither localization conditional on detection nor
successful detection-and-localization, and does not establish safe correction.

\subsection{Resources, provenance, and reproduction}

The run took 40.40 seconds on one local CPU process with one BLAS/OpenMP thread;
no GPU was used. Software was Python 3.11.2, NumPy 1.26.4, SciPy 1.13.1,
scikit-learn 1.4.2, and threadpoolctl 3.6.0. An environment-only preflight using the
old threadpoolctl 2.2.0 failed before any fit because it could not inspect the local
OpenBLAS library; that failure is retained, and only this runtime-control package
was upgraded before the scientific run. No historical experiment file was changed.

Source array SHA-256:
\begin{quote}\small\ttfamily
8b6a2145cb7523ebbae1dc2f0374b712366385bd15cca37a18a9dc6b6da0968b
\end{quote}
{\raggedright
The protocol is \path{paper/pc_validation_2026-10-17.md}; code is
\path{src/run_pc_validation_20261017.py}. Score arrays, per-episode metrics,
convergence diagnostics, environment and code/protocol hashes are in
\path{results/pc_validation_20261017/}. Input data remain subject to the original
dataset terms and are excluded from the manuscript-source ZIP.\par}

\begin{quote}\small
\begin{minipage}{\linewidth}
\begin{verbatim}
DATA=data/nvidia-libero-lerobot-v3-e590737
DATA="$DATA/libero_spatial/transitions.npz"
python src/run_pc_validation_20261017.py \
  --input "$DATA" \
  --output results/pc_validation_20261017 \
  --protocol paper/pc_validation_2026-10-17.md
\end{verbatim}
\end{minipage}
\end{quote}

The fixed four-component joint model is not refitted differently for the
conditional score. Neither favorable point estimates nor failed historical gates
are used to select a new model. The new comparison changes the interpretation of
the original result: it supports multimodal relational scoring over ridge, while
joint-density detection superiority is unsupported.

%% file: retained_numerical_details.tex
\section{Retained operating-point and ablation details}\label{sec:retained-details}

\begin{table}[H]
\caption{Locked replay evidence. Gains are absolute recall differences. ``Seeds''
are repeated episode-split seeds over fixed data, not independent trajectories.}
\label{tab:headline}
\centering
\small
\setlength{\tabcolsep}{5.2pt}
\begin{tabular}{lrrrr}
\toprule
Evaluation & Pass & Gain vs. action/range & Loc. & Max FPR\\
\midrule
Spatial, 2,000 seeds/task & 10/10 & 0.3157 & 0.8567 & 0.0279\\
Object + Goal, 500/suite & 2/2 & 0.2989 & 0.8474 & 0.0286\\
Five-family Spatial, 4,000/task & 10/10 & 0.3793 & 0.9134 & 0.0279\\
Five-family Object + Goal, 4,000/suite & 2/2 & 0.3579 & 0.8993 & 0.0284\\
Forward Spatial, fresh 2,000/task & 10/10 & 0.3739 & 0.9285 & 0.0287\\
Forward Object + Goal, fresh 2,000/suite & 2/2 & 0.3474 & 0.9092 & 0.0279\\
\bottomrule
\end{tabular}
\end{table}

\subsection{Task and suite confirmation}

Table~\ref{tab:headline} summarizes the locked feasibility evaluator. All ten
Spatial tasks pass. In the held-out LIBERO suites, relational recall is 0.3442
versus 0.0411 for the action/static comparator on Object and 0.4632 versus 0.0506
on Goal; localization is 0.9050 and 0.8936. These suites change task content but
retain the LIBERO data collection and action convention. They provide within-source
task breadth, not cross-dataset or cross-robot generalization.

The earlier six-family 2,000-seed confirmation gives hierarchical-bootstrap 95\%
intervals $[0.2805,0.3406]$ for median recall gain and $[0.8433,0.8598]$ for median
localization; maximum clean FPR lies in $[0.0272,0.0287]$. Restricting the headline
claim to five supported families increases the central values. Because that
restriction followed a family audit, we retain the six-family result and describe
the five-family result as evidence-driven claim narrowing, not a replacement of the
predeclared aggregate.

\subsection{Historical joint-mixture comparison with ridge}

The forward model has median recall 0.4504 versus 0.3560 for the joint Gaussian and
wins recall on all ten tasks, but it has lower localization on all ten: medians are
0.9285 and 0.9420. Maximum FPR is 0.0282 for forward and 0.0439 for joint. Thus the
joint comparison rejects the original notion that a conditional residual dominates
on both metrics; it reveals a calibrated recall--localization Pareto tradeoff.

Under this original ridge comparison, a deterministic four-component GMM is stronger. On 1,000 fresh split seeds, it
beats forward recall on all ten tasks by 0.15663 in the task median while keeping
maximum FPR at 0.03994. A separate localization confirmation reaches 0.95926 and
beats forward by 0.02932, with paired-task 95\% interval
$[0.02472,0.03553]$; it beats joint Gaussian by 0.01728
$[0.01398,0.01971]$. Frozen replications within Object and Goal yield recall
advantages of 0.21183 and 0.17956 and localization advantages over the better
comparator of 0.02888 and 0.02758. Again, these suites share the LIBERO convention.

Component sensitivity is not uniform. Four and eight components pass with median
recall advantages 0.16093 and 0.16341, but two components reach only 0.07252 with
interval $[0.04073,0.11486]$ and fail the frozen gate. Localization remains above
0.956. By contrast, nested calibration caps of 100, 200, and 400 clean transitions
all pass: recall advantages are 0.14526--0.15400, localization is 0.96069, and the
largest FPR is 0.04123. The evidence supports a multimodal density under modest
calibration budgets, not mixture-count invariance.

\begin{table}[H]
\caption{Paired historical evidence for the four-component GMM. Achieved FPR is not matched. The clustered rows first
average repeated split appearances within each source episode.}
\label{tab:matched}
\centering
\small
\begin{tabular}{llrr}
\toprule
Evidence & Comparison & Effect & 95\% interval\\
\midrule
Spatial splits & GMM $-$ forward recall & 0.15663 & task median\\
Spatial splits & GMM $-$ forward loc. & 0.02932 & [0.02472, 0.03553]\\
432 episodes & GMM $-$ forward recall & 0.16753 & [0.13642, 0.20297]\\
432 episodes & GMM $-$ forward loc. & 0.03089 & [0.02679, 0.03541]\\
432 episodes & GMM $-$ joint loc. & 0.01837 & [0.01522, 0.02198]\\
Metadata baseline & metadata $-$ GMM recall & -0.31196 & [-0.33328, -0.24729]\\
Metadata baseline & metadata $-$ GMM loc. & -0.10914 & [-0.12655, -0.09866]\\
\bottomrule
\end{tabular}
\end{table}

\subsection{Episode-clustered effect and matched static comparator}

The fresh clustered analysis uses 100 new partitions per Spatial task but collapses
repeated appearances within 432 immutable task--episode clusters before inference.
The equal-task/equal-episode GMM-minus-forward recall effect is 0.16753 with paired
95\% interval $[0.13642,0.20297]$. Localization effects are 0.03089 over forward
and 0.01837 over joint Gaussian. Exact task-sign-flip $p=1/1024$ for each direction;
Holm-adjusted $p=0.00293$ across the frozen three-comparison family. All 1,000 GMM
fits converge, maximum task FPR is 0.0444, and every episode appears in at least 11
test splits. This analysis supports the Spatial ordering without treating split
runs as independent trajectories. An unchanged Object/Goal extension crashes
because one Object task has no valid gripper-polarity rows; we make no clustered
claim for those suites.

The field-aligned displacement contract is non-degenerate: task-median recall is
0.32020, localization 0.84245, and maximum FPR 0.02983. Nevertheless, it passes
0/10 GMM-noninferiority gates. Metadata-minus-GMM recall is $-0.31196$
$[-0.33328,-0.24729]$, and localization is $-0.10914$
$[-0.12655,-0.09866]$. This comparison rejects a range-check-only explanation,
while its missing controller specification prevents us from presenting it as a
faithful third-party implementation.

\subsection{Retained numerical ablations and evidence hierarchy}

\paragraph{Which relation matters?}
The earliest detector combined a forward residual with an inverse model that
predicts $a_t$ from $(s_t,s_{t+1})$. On 500 unseen split seeds per Spatial task,
forward-only, inverse-only, and combined variants all pass the action/static gate.
The combined score raises paired median recall by only 0.0065 over forward, with a
task-bootstrap 95\% interval $[-0.0042,0.0282]$ that includes zero. Forward-only
localization is higher on all ten tasks: combined-minus-forward has paired median
$-0.0053$, interval $[-0.0094,-0.0026]$. The original study therefore retained forward-only as its
parsimonious ridge conditional baseline. This ablation does not privilege a deterministic
mechanism, because the later GMM comparison shows an advantage over ridge under the
original calibration contract. The post-decision conditional comparison does not
support joint-density superiority.

The original methods share a calibration quantile and a 5\% test-FPR ceiling,
but their achieved operating points differ. Forward and joint-Gaussian maximum
clean FPRs are 0.0282 and 0.0439; forward and mixture maxima in the passing
confirmation are 0.02763 and 0.03994. These ceilings do not establish a matched-FPR
advantage. Section~\ref{sec:pc-validation} supplies that comparison and the
capacity-controlled conditional baseline. The earlier 200-split mixture screen
reaches 0.0500375 and remains inconclusive under its frozen gate.

\paragraph{Family and severity sensitivity.}
The forward-versus-joint tradeoff is transformation-dependent. Forward is better on
both recall and localization for scale, with +0.2620 recall and +0.0098
localization; joint is better on both for X-sign, with +0.0242 recall and +0.0277
localization. Only rotation shrink satisfies the frozen material Pareto gate. A
paired task bootstrap makes the limits sharper: scale recall remains directional at
$[0.2263,0.3125]$, but its localization interval $[-0.0181,0.0019]$ crosses zero;
for X-sign, joint localization remains directional at $[0.0207,0.0319]$, while the
recall interval crosses zero. Aggregate detector rankings therefore should not be
read as uniform family dominance.

Fault magnitude changes explanation before it changes detection. For X offsets of
0.01, 0.025, 0.05, 0.10, and 0.20, GMM recall is
0.0305/0.0314/0.0330/0.0424/0.1518 while exact-magnitude localization rises from
0.4710 to 0.9490. The curve is monotone but was run after the failed offset screen,
so it is mechanism evidence rather than a rescued supported family. Temporal
severity behaves similarly: forward-minus-action recall gain rises from
0.0049/0.0359 at one/two frames to 0.2926/0.7179/0.9209 at
four/eight/sixteen frames. Severe misalignment becomes detectable, while the short,
smoother latency most relevant to deployment remains unsupported. A learned
temporal baseline using $(s_t,a_{t-1})$ also passes 0/10 short-temporal gates, which
rules out only that simple feature omission, not richer sequence monitoring.

\paragraph{Task breadth and finite-source limits.}
Task-held-out triplets pass 3/3 with median recall gain 0.1740, localization 0.7760,
and maximum FPR 0.0143. Object and Goal repeats strengthen within-LIBERO task
breadth. They do not change the action representation or data-collection convention.
For the same reason, 4,000 deterministic split seeds cannot substitute for 4,000
new episodes. The 432-episode clustered estimand was the original headline interval because it
makes that dependency explicit. The attempted all-task clustered extension is
retained as a protocol failure rather than silently changing its fault vocabulary
when one Object task has no valid gripper-polarity rows.

We use an evidence hierarchy to prevent reuse from widening the claim. Fresh splits
under the frozen evaluator establish algorithmic ordering on recorded tasks.
Episode-clustered analyses change the inferential unit and support only the Spatial
effect for which the gate is well-defined. Selection-frozen reuses diagnose
heterogeneity, false negatives, and candidate confusion. Post-hoc severity curves
explain a failed family. External screens test whether the same contract survives a
new convention. These levels answer different questions; none converts a
descriptive mechanism audit into an independent robot or task replication.